\documentclass[10pt, a4paper]{article}

\usepackage[]{lrec-coling2024} 
\usepackage{times}
\usepackage{latexsym}
\usepackage{algorithm}
\usepackage{algorithmic}
\usepackage{multirow}
\usepackage{multicol}
\usepackage{graphicx}
\usepackage{amssymb}
\usepackage{amsmath}
\usepackage{tabularx}

\title{Transformer-based Joint Modelling for Automatic Essay Scoring and Off-Topic Detection} 

\name{Sourya Dipta Das$^*$, Yash Vadi$^*$, Kuldeep Yadav 
} 

\address{SHL Labs, India \\
         sourya.das@shl.com, yash.vadi@yahoo.com, kuldeep.yadav@shl.com\\
}

\abstract{
Automated Essay Scoring (AES) systems are widely popular in the market as they constitute a cost-effective and time-effective option for grading systems. Nevertheless, many studies have demonstrated that the AES system fails to assign lower grades to irrelevant responses. Thus, detecting the off-topic response in automated essay scoring is crucial in practical tasks where candidates write unrelated text responses to the given task in the question. In this paper, we are proposing an unsupervised technique that jointly scores essays and detects off-topic essays. The proposed Automated Open Essay Scoring (AOES) model uses a novel topic regularization module (TRM), which can be attached on top of a transformer model, and is trained using a proposed hybrid loss function. After training, the AOES model is further used to calculate the Mahalanobis distance score for off-topic essay detection. Our proposed method outperforms the baseline we created and earlier conventional methods on two essay-scoring datasets in off-topic detection as well as on-topic scoring. Experimental evaluation results on different adversarial strategies also show how the suggested method is robust for detecting possible human-level perturbations. 
 \\ \newline  }

\begin{document}

\maketitleabstract

\def\thefootnote{*}\footnotetext{These authors contributed equally to this work}

\section{Introduction}
Writing assessments are used in many business and academic institutions to measure the written language competency of prospective employees or students, and AES are often widely used to automate the grading process. The candidates are assessed based on the written essay by taking multiple factors into consideration, such as grammar usage, choice of word style to convey the central idea, ability to write a coherent piece of text, factuality, relevance, etc. In spite of many deep learning and transformer-based methods \citet{yang2020enhancing,wang2022use} showing high human-level agreement scores with these AES systems, ~\citet{kabra2022evaluation,parekh2020my,ding2020don,perelman2020babel} have showcased that many automated scoring systems are vulnerable to an adversarial attack by the test-taker. Specifically, ~\citet{kabra2022evaluation} has showcased that different state-of-the-art AES methods suffer from adversarial responses and fail to provide a low score for them. Moreover, we found that adding unrelated content improved the scores. ~\citet{parekh2020my} has showcased that AES is \textit{overstable} (large change in input essay response but little or no change in output score). For instance, some candidates could attempt to write a planned response that is unrelated to the question in an effort to inflate their score. These unrelated responses are not related to the question prompt and should not be graded more than zero on the content score. It is crucial to develop an efficient assessment scoring system that can flag these responses in order to validate the assessment scores and maintain trustworthiness.

In the real scenario, these off-topic responses might arise from a wide variety of sources. Furthermore, using a supervised approach for training a model to classify whether the response is on-topic or off-topic will not generalize well ~\citet{xu2021unsupervised}, as collecting off-topic responses with every possible combination is not practically possible. 
Based on the success of the transformer-based model~\citet{yang2020enhancing,wang2022use,ludwig2021automated} in natural language understanding, we used the BERT~\citet{devlin2018bert} model for this study. In this paper, we present an approach that can be jointly used for essay grading and off-topic detection. We propose the AOES model with an additional regularization branch, that calibrates the regression output of the model. We further showcased that the proposed architecture with Mahalanobis distance can be utilized for both essay scoring and off-topic essay detection. Additional testing on the adversarial test cases demonstrates that this approach is immune to detecting adversarial responses. So, the proposed model offers a useful compromise whereby humans just need to assess a few samples that have been indicated by the detector models in suspicion of cheating or mischievous activity. Our contributions in this paper are listed below:

\begin{itemize}
    \item We propose a novel multi-task joint AOES model that can be used to jointly score the on-topic essay and detect off-topic responses, unlike previous methods where separate models are used for essay score estimation and off-topic text detection.
    \item We present a Mahalanobis distance-based unsupervised approach for off-topic detection that does not require additional off-topic data during training.
    \item We evaluate our method on two essay datasets, ASAP-AES, an open source dataset, and PsyW-Essay, an in-house industrial dataset, and have also shown that AOES can consistently improve upon baseline methods and previous supervised, unsupervised state-of-the-art methods.
    \item We also evaluate our method on various off-topic adversarial perturbations and show effectiveness in the detection of these adversarial samples.
\end{itemize}

\section{Related Work}
In recent years, there has been some research work on off-topic detection. Off-topic Detection has been explored in both supervised and unsupervised settings for both essay and transcribed spoken responses. There are very few works done in unsupervised settings particularly. 
We begin with an unsupervised method ~\citet{louis2010off} who proposed two methods that expand the short question prompt with the words most likely to appear in the essay with respect to that prompt after applying spelling correction in the response text. After that, they compared the similarity between the response essay and corresponding question prompt to detect the off-topic essay.
In supervised methods, ~\citet{wang2019automatic} suggested a method that first creates a similarity grid for each pair of responses and its corresponding question prompt. This similarity grid will be then fed into the Inception net to classify whether the response belongs to that prompt or not. ~\citet{shahzad2022computerization} proposed a method by combining idf weighted word, average word embeddings, and word mover distance embedding vectors together and then trained a random forest classifier for detection. ~\citet{yoon2017off} proposed an automatic filtering model that uses both a set of linguistic features like vocabulary, and grammar skills and document semantic similarity features based on word hypotheses and content models to detect off-topic responses. A subset of the features listed was also leveraged by other studies, including ~\citet{huang2018off} and ~\citet{lee2017off}, to access similarity between questions and responses, and these features were subsequently used to train deep networks. ~\citet{raina2020complementary} combined Hierarchical attention based topic model (HATM) and Similarity Grid model (SGM) for off-topic spoken essay detection. ~\citet{malinin2016off} proposed a Question Topic Adaptive RNNLM framework that learns to associate candidate responses to given questions with samples in a topic space constructed using these responses only. 
But According to the latest study~\citet{singla2021aes}, most of these previous off-topic detection models cannot detect adversarial samples.


\section{Problem Statement}
In the Automatic Essay Scoring System, a candidate's written essay will be either an on-topic essay response, which will be evaluated by the system or an off-topic essay response, which will be rejected by the system and given zero score. This problem statement is formally defined below. 

We are given an on-topic essay training set $S_{train}=\{\mathrm{X_{e}}, \mathrm{Y_g}\}=\left\{\left(X_{e}^i, Y_{g}^i\right)\right\}_{i=1}^M$, where $M$ is number of training samples. Each input sample $\left(X_{e}^i, Y_{g}^i\right)$, an candidate's written response text $X_{e}^i$ and its assessment score $Y_{g}^i \in\mathbb{Q}$. 
During inference, test-set, $S_{test} = \{\mathrm{\hat{X}_{e}}, \mathrm{\hat{Y}_g}, \mathrm{C_g}\}=\left\{\left(\hat{X}_{e}^i, \hat{Y}_{g}^i, C_{g}^i\right)\right\}_{i=1}^K$, where $K$ is number of samples in test-set. 
Each input sample $\left(\hat{X}_{e}^i, \hat{Y}_{g}^i, C_{g}^i\right)$, additionally has essay type class label $C_{g}^i \in\ \{C_{on}, C_{off}\}$, where, $C_{on}, C_{off}$ are class label ids for on-topic, off-topic essays respectively. We evaluate our model on this test set. Our goal is to train a joint model only on
on-topic essay training data, $S_{train}$ such that the proposed model is able to: 1) Correctly predict whether the essay is on-topic or not. 2) Estimate on-topic essay scores precisely or flag the off-topic response and give them zero score. The proposed model can be
described as follows:
\begin{equation}
Y^i_s, Y^i_p = \begin{cases} F_E(\hat{X}^i_{e}), C_{on} & \text { if } D^t_{MD}(\hat{X}^i_{e}) \leq \delta \\ 0, C_{off} & \text { if } D^t_{MD}(\hat{X}^i_{e}) > \delta\end{cases}
\end{equation}
where $Y^i_s, Y^i_p$  are the predicted essay score and predicted essay type class from the proposed model, $F_E(\hat{X}^i_{e})$ for $i$-th test-set sample respectively, $D^t_{MD}(\hat{X}^i_{e})$ is an off-topic score estimation function that determines if the input corresponds to the on-topic or off-topic class and $\delta$ is a threshold value. It should be noted that our system assigns zero as an essay assessment score to all detected off-topic essays.

\section{Proposed Methodology}
Our proposed method takes advantage of the training data coming mainly from on-topic text data by using multi-task learning. We use an additional regularization along with regression loss to place a constraint on the final prediction score. This extra regularization is introduced with the aim of better performance on the on-topic essay scoring. We make use of this fact and provide a Mahalanobis distance-based method for a transformer-based model to detect off-topic text since the improved performance is due to a more valid and reliable feature representation ~\citet{hsu2020generalized}. We refer to our proposed method as Automated Open Essay Scoring (AOES) System in this paper.
\subsection{Model Architecture}
\begin{figure}[t]
  \centering
  \includegraphics[scale=0.45]{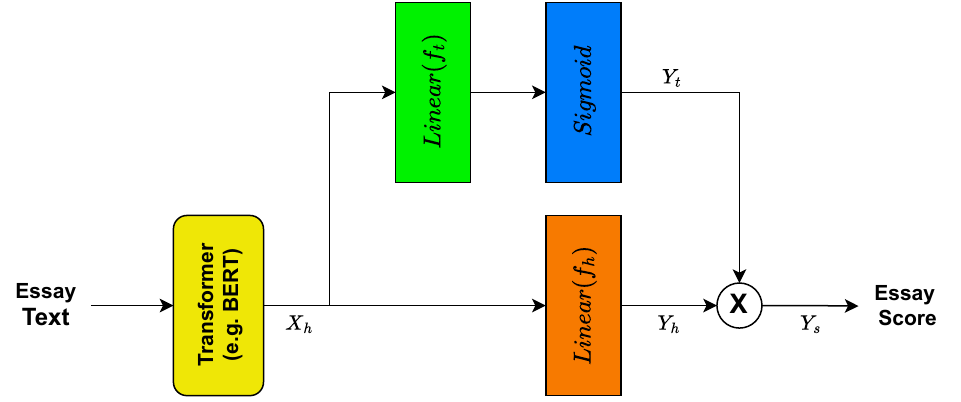} 
  \caption{Proposed AOES Model Architecture Diagram}
  \label{fig:AOES}
\end{figure}

We have used a pre-trained BERT~\citet{devlin2018bert}, a transformer-based model as the backbone, and a Topic Regularization Module (TRM) layer which is used like a simple drop-in replacement of the linear regression layer. The whole model architecture is illustrated in Figure~\ref{fig:AOES}. Further details of the model and TRM layer are discussed in the following section.
\subsection{Topic Regularization Module (TRM)}
We design the TRM to mitigate the overestimation of the regression score by decomposing the final score into two separate branches as shown in Figure~\ref{fig:AOES}. The lower branch is the main regression scoring branch where BERT hidden state pooled output, $X_h$ is passed through a normal linear layer represented as a function, $f_h$ to return a non-calibrated score, $Y_h$ as mentioned in Equation~\ref{Yh_eqn}. The upper branch is responsible for calibrating the initial regression score from $f_h$ to compensate for the overestimated regression score. It uses another linear layer represented as a function, $f_t$ to generate a scaling factor score, $Y_t$ from the same BERT hidden state pooled output, $X_h$. Later, final regression score, $Y_s$ as essay score is enumerated by multiplying both $Y_h$ and $Y_t$ as mentioned in Equation~\ref{TRM_eqn}. 
\begin{gather} 
Y_h = f_h(X_h)\, \label{Yh_eqn} \\ 
Y_t = Sigmoid(f_t(X_h)) \\
Y_s = Y_t*Y_h  \label{TRM_eqn}
\end{gather}
\subsection{TRM Training Loss Function} \label{training_loss_func}
AOES model is trained using a hybrid loss function, $\mathcal{L}_{hybrid}$ as mentioned in Equation~\ref{loss_eqn}. This hybrid loss function consists of two other loss functions which are mean square loss, $\mathcal{L}_{MSE}$ and Topic Regularization Loss, $\mathcal{L}_{Topic}$. The $\mathcal{L}_{MSE}$ aims to minimize the mean square error between the final predicted score, $Y_s$ and actual graded essay score, $Y_g$.  The $\mathcal{L}_{Topic}$ aims to calibrate the initial regression score, $Y_h$ to the final regression score, $Y_s$ such that it also aids in minimizing mean square loss, $\mathcal{L}_{MSE}$. The output, $Y_t$ is restricted between 0 and 1. Then, this loss function, $\mathcal{L}_{Topic}$ encourages $Y_t$ close to 1 for on-topic training data samples.
\begin{gather} \label{loss_eqn}
\mathcal{L}_{hybrid} = \mathcal{L}_{MSE} + \lambda\mathcal{L}_{Topic}  \\
\mathcal{L}_{MSE} = \frac{1}{N}\sum_{i=1}^{N}||Y^i_g - Y^i_s||^2 \label{equ:mse_loss} \\
\mathcal{L}_{Topic} = -\frac{1}{N}\sum_{i=1}^{N}log(Y^i_t)
\end{gather}
Here, $N$ is the number of samples in the batch.
$\mathcal{L}_{Topic}$ is used to incorporate extra regularization to attenuate the initial overestimated regression score for training data samples. It is important to note that $Y_t$ is not used to directly predict if the input text sample is off-topic or not.

\subsection{Off Topic Detection Method}
For off-topic response detection, we have used latent feature-based Mahalanobis distance score as off-topic detection score. This Mahalanobis distance score is calculated by using latent representations from all layers of the finetuned AOES Model $F_E(x)$, inspired by the previous work~\citet{xu2021unsupervised}. As explored in work~\citet{jawahar2019does}, latent feature vector from different layers of the transformer model is used to capture different aspects of language, such as lower layers that capture lexical features, middle layers that represent syntactic features, and higher layers that encode semantic properties.

Training samples, $X_e$ are fed into the fine-tuned model, $F_E(.)$ to extract intermediate layer embeddings and then apply a hidden layer activation function, $tanh(.)$ to transform the previous intermediate layer features into latent embedding feature vectors, i.e. $[h^1_i, h^2_i,\dots, h^L_i] \in \mathbb{R}_{d.L}$ where $d$ is the dimension of embedding vector and L is the number of intermediate layers. Then, mean and covariance of training data, $S_{train}$ are estimated by the following equations.
\begin{gather}
\tiny
    \mu_l=\frac{1}{M} \sum_{i=1}^{M}h^l_i \\
    \Sigma_{l} =\frac{1}{M} \sum_{i=1}^{M}\left(h^{l}_i-\mu_l\right)\left(h^{l}_i-\mu_l\right)^{T} 
\end{gather}
where, for $l$-th layer of the model, $h^l_i$ is extracted latent feature vector of $i$-th training data sample and $\mu_l, \Sigma_l$ are corresponding means, covariances of the feature vectors from all $M$ training data samples. 
The calculated $\mu_l$ and $\Sigma_l$ are further used to calculate Mahalanobis distance at the inference time on the test set, $S_{test}$. $D^l_j$ is the $l$-th layer Mahalanobis distance of $j$-th test data sample during inference.
Now, the Mahalanobis distance score, $D^t_j$ is calculated by summing layer-wise Mahalanobis distances up across all layers for the $j$-th test data sample. This Mahalanobis distance score, $D^t_j$ is applied as the output of the off-topic score estimation function,$ D^t_{MD}(\hat{X}^j_{e})$ with a threshold value, $\delta$ for off-topic essay detection.
\begin{gather}
    D^l_j =\left(h^{l}_j-\mu_l\right)^T\Sigma_l^{-1}\left(h^{l}_j-\mu_l\right)  \label{md_dist} \\
    D^t_j = \sum_{l=1}^{L}D^l_j
\end{gather}
\begin{figure*}[t]
  \centering
  \includegraphics[scale=0.6]{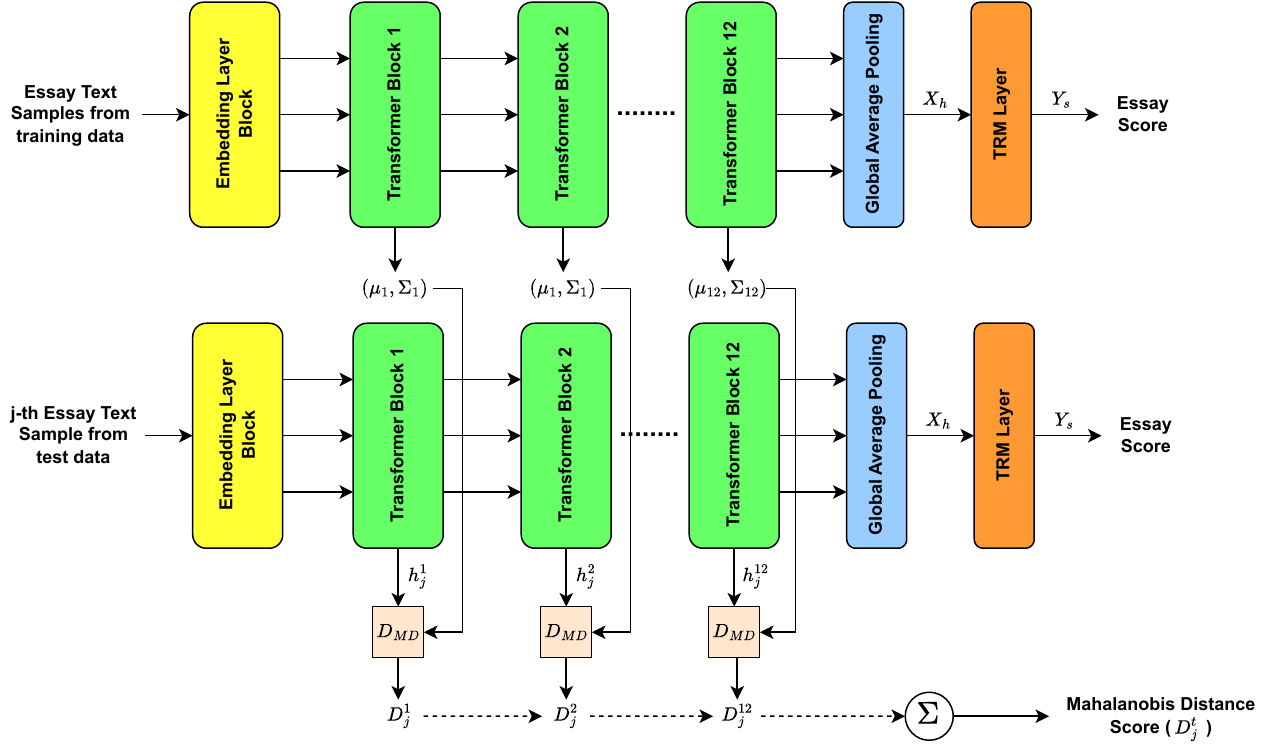}
  \caption{A Overview of Off-Topic Detection Process Diagram. Here, the $D^t_{MD}$ function is used to calculate layer-wise Mahalanobis distance as mentioned in Equation~\ref{md_dist}.  }
  \label{fig:Offtopic_AOES}
\end{figure*}

\section{Experiments and Results}

\subsection{System Description}
We have implemented our model using PyTorch and pre-trained BERT base model from Hugging Face transformer library~\footnote{https://huggingface.co}. We train our model and baselines on a machine with Intel Xeon Platinum 8124M CPU, 16GB RAM, and one 12 GB NVIDIA GTX 1080 GPU. We fine-tuned it with the same hyperparameters from the original model. We have used both Scikit-Learn~\footnote{https://scikit-learn.org}, Scipy~\footnote{https://scipy.org/} python packages for evaluation purposes. 

\subsection{Dataset Details}

\textbf{ASAP-AES Dataset : }
The ASAP-AES dataset~\footnote{https://osf.io/9fdrw/} contains data from 8 different essay prompts with 3 distinctive types of essays. The ASAP-AES dataset has argumentative essays, response essays, and narrative essays. All the essays were written by native English-speaking children from classes 7 to 10. For our experiments, we have utilized all 8 prompts and have used 20 percent of the data as a test set for each prompt. Further information about the dataset is provided in Table~\ref{tab:asap_tab}. 

\begin{table}
\scriptsize
\centering
\begin{tabular}{|c|c|c|c|c|}  
\hline
\begin{tabular}[c]{@{}c@{}}\textbf{Prompt}\\\textbf{ID}\\\end{tabular} & \begin{tabular}[c]{@{}c@{}}\textbf{Essay }\\\textbf{Type}\end{tabular} & \begin{tabular}[c]{@{}c@{}}\textbf{Average~ }\\\textbf{Length}\end{tabular} & \begin{tabular}[c]{@{}c@{}}\textbf{No. of }\\\textbf{Essays}\end{tabular} & \begin{tabular}[c]{@{}c@{}}\textbf{Score }\\\textbf{Range}\end{tabular} \\ 
\hline
1 & Argumentative & 350 & 1783 & 2 - 13 \\ 
\hline
2 & Argumentative & 350 & 1800 & 1 - 6 \\ 
\hline
3 & Source Dependent & 150 & 1726 & 0 - 3 \\ 
\hline
4 & Source Dependent & 150 & 1772 & 0 - 3 \\ 
\hline
5 & Source Dependent & 150 & 1805 & 0 - 4 \\ 
\hline
6 & Source Dependent & 150 & 1800 & 0 - 4 \\ 
\hline
7 & Narrative & 300 & 1569 & 0 - 30 \\ 
\hline
8 & Narrative & 650 & 723 & 0 - 60 \\
\hline
\end{tabular}
\caption{ASAP-AES Dataset Details and Statistics}
\label{tab:asap_tab}
\end{table}

\textbf{PsyW-Essay Dataset : }
PsyW-Essay Dataset is created from a product that is an online psychometric assessment designed to assess an individual’s ability to write effectively in the English language. In this test, the candidate is supposed to write an essay on the provided topic. It also takes into consideration content-related aspects such as the candidate's view on the topic, how relevant the essay is to the given topic, and how the candidates organize their own flow of thoughts.
The current dataset subset, which consists of 22 separate prompts, is used to assess test takers' writing proficiency. To determine the final score, the dataset was rated by both expert rater and a group of raters.
 Here, we selected 9 prompts for the experiments and used 20 percent of the data as test set for each prompt. Table~\ref{tab:PsyW-Essay_tab} contains information about the dataset.

\begin{table}[]
\scriptsize
\centering
\begin{tabular}{|c|c|c|c|} 
\hline
\begin{tabular}[c]{@{}c@{}}\textbf{Prompt}\\\textbf{ID}\end{tabular} & \begin{tabular}[c]{@{}c@{}}\textbf{Average}\\\textbf{Length}\end{tabular} & \begin{tabular}[c]{@{}c@{}}\textbf{No. of }\\\textbf{Essays}\end{tabular} & \begin{tabular}[c]{@{}c@{}}\textbf{Score }\\\textbf{Range}\end{tabular} \\ 
\hline
1 & 224 & 675 & 0-5 \\ 
\hline
2 & 236 & 699 & 0-5 \\ 
\hline
3 & 251 & 732 & 0-5 \\ 
\hline
4 & 235 & 727 & 0-5 \\ 
\hline
5 & 237 & 682 & 0-5 \\ 
\hline
6 & 237 & 721 & 0-5 \\ 
\hline
7 & 240 & 709 & 0-5 \\ 
\hline
8 & 223 & 667 & 0-5 \\ 
\hline
9 & 232 & 692 & 0-5 \\
\hline
\end{tabular}
\caption{PsyW-Essay Dataset Details and Statistics}
\label{tab:PsyW-Essay_tab}
\end{table}

\begin{table*}[t]
\scriptsize
\centering
\begin{tabular}{|c|c|c|c|c|c|c|c|c|c|c|c|} 
\hline
\multirow{2}{*}{\textbf{Dataset}}    & \multirow{2}{*}{\textbf{Model}}    & \multirow{2}{*}{\textbf{Metric}} & \multicolumn{9}{c|}{\textbf{Prompt ID }}                                                                                                                \\ 
\cline{4-12}
                                     &                                    &                                  & \textbf{1}     & \textbf{2}     & \textbf{3}     & \textbf{4}     & \textbf{5}     & \textbf{6}     & \textbf{7}     & \textbf{8}     & \textbf{9}      \\ 
\hline
\multirow{8}{*}{PsyW-Essay} & \multirow{2}{*}{Baseline 1} & QWK                              & 0.698          & 0.720          & 0.723          & 0.796          & 0.733          & 0.713          & 0.752          & 0.710          & 0.709           \\
                                     &                                    & Correlation                      & 0.781          & 0.786          & 0.873          & 0.890          & 0.791          & 0.753          & 0.876          & 0.799          & 0.786           \\ 
\cline{2-12}
                                     & \multirow{2}{*}{Baseline 2}     & QWK                              & 0.703 & 0.728 & 0.718 & 0.812 & 0.513 & 0.688 & 0.765 & 0.673 & 0.730  \\
                                     &                                    & Correlation                      & 0.786 & 0.715 & 0.875 & 0.902 & 0.714 & 0.807 & 0.872 & 0.731 & 0.808  \\
\cline{2-12}
                                     & \multirow{2}{*}{AOES (BERT)}     & QWK                              & \textbf{0.722 }& \textbf{0.734} & 0.733 & 0.826 & \textbf{0.742} & \textbf{0.720} & \textbf{0.807} & \textbf{0.728} & \textbf{0.763}  \\
                                     &                                    & Correlation                      & \textbf{0.809} & \textbf{0.797} & \textbf{0.899} & \textbf{0.906} & \textbf{0.814} & \textbf{0.822} & 0.878 & \textbf{0.803} & \textbf{0.824}  \\
\cline{2-12}
                                     & \multirow{2}{*}{AOES (RoBERTa)}     & QWK                              & 0.681 & 0.629 & \textbf{0.757} & \textbf{0.832} & 0.694 & 0.682 & 0.778 & 0.563 & 0.651  \\
                                     &                                    & Correlation                      & 0.777 & 0.755 & 0.883 & 0.893 & 0.805 & 0.735 & \textbf{0.879} & 0.796 & 0.780  \\
\hline
\multirow{8}{*}{ASAP-AES} & \multirow{2}{*}{Baseline 1} & QWK                              & 0.784    & 0.647     & 0.611     & 0.784     & 0.741     & 0.765     & 0.839     & 0.749     & -          \\
                                   &                                    & Correlation                      & 0.832      & 0.731     & 0.685     & 0.822     & 0.817     & 0.828     & 0.855     & 0.737 & -          \\ 
\cline{2-12}
                                   & \multirow{2}{*}{Baseline 2}    & QWK                              & 0.777 & 0.636 & 0.540 & 0.772 & 0.760 & 0.764 & 0.829 & 0.598 & -  \\
                                   &                                    & Correlation                      & 0.818 & 0.739 & 0.677 & 0.822 & 0.812 & 0.824 & 0.859 & 0.658 & -  \\
\cline{2-12}
                                   & \multirow{2}{*}{AOES (BERT) }    & QWK                              & \textbf{0.793}  & 0.661 & \textbf{0.667} & 0.789 & \textbf{0.782}  &  \textbf{0.787} & \textbf{0.845} & \textbf{0.754} & -  \\
                                   &                                    & Correlation                      & \textbf{0.834}     & 0.741 & 0.695 & 0.836  & 0.825 & \textbf{0.842}  & \textbf{0.863} & \textbf{0.789}& -  \\
\cline{2-12}
                                   & \multirow{2}{*}{AOES (RoBERTa)}    & QWK                              & 0.785 & \textbf{0.680} & 0.635 & \textbf{0.822} & 0.705 & 0.781 & 0.842 & 0.734 & -  \\
                                   &                                    & Correlation                      & 0.827 & \textbf{0.746} & \textbf{0.698} & \textbf{0.846} & \textbf{0.828} & 0.833 & 0.854 & 0.766 & -  \\
\hline
\end{tabular}
\caption{On-Topic Automatic Essay Scoring Quantitative Comparison Results on Both Datasets}
\label{tab:PsyW_asap_ontopic}
\end{table*}

\begin{table*}
\scriptsize
\centering
\begin{tabular}{|c|c|c|c|c|c|c|c|c|c|c|c|} 
\hline
\multirow{2}{*}{\textbf{Dataset}}     & \multirow{2}{*}{\textbf{Model}}               & \multirow{2}{*}{\textbf{Metric}} & \multicolumn{9}{c|}{\textbf{Prompt ID }}                                                                                                                \\ 
\cline{4-12}
                                      &                                               &                                  & \textbf{1}     & \textbf{2}     & \textbf{3}     & \textbf{4}     & \textbf{5}     & \textbf{6}     & \textbf{7}     & \textbf{8}     & \textbf{9}      \\ 
\hline
\multirow{9}{*}{PsyW-Essay } & \multirow{3}{*}{AOES (without TRM)}  & F1                               & 0.602          & 0.920          & 0.874          & 0.847          & 0.758          & 0.645          & 0.613          & 0.787          & 0.721           \\
                                      &                                               & Precision                        & 0.985          & 0.991          & 0.973          & 0.998          & 0.998          & 0.998          & 0.998          & 0.998          & 0.976           \\
                                      &                                               & Recall                           & 0.433          & 0.858          & 0.794          & 0.734          & 0.610          & 0.476          & 0.442          & 0.648          & 0.571           \\ 
\cline{2-12}
                                      & \multirow{3}{*}{AOES (with L2 Loss)} & F1                               & 0.827          & 0.855          & 0.892          & 0.915          & 0.859          & 0.814          & 0.866          & 0.853          & 0.785           \\
                                      &                                               & Precision                        & 0.841          & 0.852          & 0.897          & 0.922          & 0.902          & 0.825          & 0.897          & 0.878          & 0.838           \\
                                      &                                               & Recall                           & 0.813          & 0.858          & 0.890          & 0.909          & 0.822          & 0.803          & 0.838          & 0.830          & 0.738           \\ 
\cline{2-12}
                                      & \multirow{3}{*}{AOES (BERT)}                & F1                               & \textbf{0.899} & \textbf{0.929} & \textbf{0.894} & \textbf{0.919} & \textbf{0.887} & \textbf{0.869} & \textbf{0.888} & \textbf{0.884} & \textbf{0.804}  \\
                                      &                                               & Precision                        & 0.911          & 0.929          & 0.902          & 0.928          & 0.913          & 0.881          & 0.914          & 0.905          & 0.851           \\
                                      &                                               & Recall                           & 0.887          & 0.929          & 0.886          & 0.909          & 0.864          & 0.857          & 0.863          & 0.864          & 0.762           \\
\hline
\hline
\multirow{9}{*}{ASAP-AES} & \multirow{3}{*}{AOES (without TRM) } & F1                               & 0.865          & 0.887          & 0.833          & 0.669          & 0.896          & 0.952          & 0.613          & 0.934     & -      \\
                                   &                                               & Precision                        & 0.809          & 0.840          & 0.772          & 0.611          & 0.846          & 0.931          & 0.514          & 0.922     & -      \\
                                   &                                               & Recall                           & 0.930          & 0.940          & 0.905          & 0.738          & 0.952          & 0.973          & 0.760          & 0.950      & -     \\ 
\cline{2-12}
                                   & \multirow{3}{*}{AOES (with L2 Loss)} & F1                               & 0.896          & 0.868          & 0.875          & 0.644          & 0.891          & \textbf{0.956} & 0.775          & 0.945     & -      \\
                                   &                                               & Precision                        & 0.848          & 0.821          & 0.824          & 0.584          & 0.845          & 0.939          & 0.693          & 0.941       & -    \\
                                   &                                               & Recall                           & 0.950          & 0.920          & 0.933          & 0.719          & 0.942          & 0.972          & 0.880          & 0.950     & -      \\ 
\cline{2-12}
                                   & \multirow{3}{*}{AOES (BERT)}               & F1                               & \textbf{0.900} & \textbf{0.891} & \textbf{0.883} & \textbf{0.782} & \textbf{0.908} & 0.954 & \textbf{0.801} & \textbf{0.970} & - \\
                                   &                                               & Precision                        & 0.856          & 0.847          & 0.838          & 0.739          & 0.868          & 0.956          & 0.744          & 0.970      & -     \\
                                   &                                               & Recall                           & 0.950          & 0.940          & 0.933          & 0.831          & 0.951          & 0.972          & 0.870          & 0.970    & -       \\
\hline
\end{tabular}
\caption{Off-Topic Detection Results of the Ablation Study on Both ASAP-AES and PsyW-Essay Datasets}
\label{tab:PsyW_asap_offtopic_ablation}
\end{table*}

\begin{table*}[h]
\scriptsize
\centering
\begin{tabular}{|c|c|c|c|c|c|c|c|c|c|} 
\hline
\multirow{2}{*}{\textbf{Model}} & \multirow{2}{*}{\textbf{Metric}} & \multicolumn{8}{c|}{\textbf{Prompt ID }} \\ 
\cline{3-10}
 &  & \textbf{1} & \textbf{2} & \textbf{3} & \textbf{4} & \textbf{5} & \textbf{6} & \textbf{7} & \textbf{8} \\ 
\hline
\multirow{3}{*}{Baseline 1} & F1 & 0.507 & 0.347 & 0.616 & 0.630 & 0.342 & 0.939 & 0.198 & 0.734 \\ 
& Precision & 1.000 & 1.000 & 0.978 & 0.630 & 0.222 & 0.949 & 1.000 & 1.000 \\ 
& Recall & 0.340 & 0.210 & 0.450 & 0.630 & 0.740 & 0.930 & 0.110 & 0.580 \\ 
\hline
\multirow{3}{*}{Baseline 2} & F1 & 0.853 &  0.884 &  0.809 &  0.777 &  0.734 &  0.942 &  0.791 &  0.798 \\ 
& Precision & 0.990  &  0.929  &  0.949  &  0.664  &  0.954  &  0.951  &  0.946  &  0.998 \\ 
& Recall & 0.75  &  0.843  &  0.705  &  0.937  &  0.596  &  0.934  &  0.68  &  0.665 \\ 
\hline
\multirow{3}{*}{\citet{louis2010off}}  & F1 & 0.669 & 0.644 & 0.350 & 0.586 & 0.328 & 0.751 & 0.284 & 0.806 \\ 
 & Precision & 0.566 & 0.545 & 0.260 & 0.522 & 0.240 & 0.669 & 0.207 & 0.783 \\ 
 & Recall & 0.820 & 0.790 & 0.533 & 0.669 & 0.519 & 0.856 & 0.450 & 0.830 \\ 
\hline
\multirow{3}{*}{\citet{shahzad2022computerization}}  & F1 & 0.888 & 0.854 & 0.336 & 0.620 & 0.131 & 0.889 & 0.279 & 0.608 \\ 
 & Precision & 0.954 & 0.929 & 0.719 & 0.746 & 0.444 & 0.958 & 0.528 & 0.732 \\ 
 & Recall & 0.830 & 0.790 & 0.219 & 0.531 & 0.077 & 0.829 & 0.190 & 0.520 \\ 
\hline
\multirow{3}{*}{\citet{raina2020complementary} }  & F1 & 0.851 & 0.849 & 0.721 & 0.776 & 0.673 & 0.881 & 0.713 & 0.800 \\ 
 & Precision & 0.934 & 0.886 & 0.758 & 0.846 & 0.665 & 0.941 & 0.698 & 0.785 \\ 
 & Recall & 0.783 & 0.815 & 0.689 & 0.718 & 0.682 & 0.829 & 0.729 & 0.816 \\ 
\hline
\multirow{3}{*}{AOES (BERT) } & F1 & 0.900 & 0.891 & 0.883 & \textbf{0.782} & \textbf{0.908} & \textbf{0.954} & \textbf{0.801} & \textbf{0.970} \\
& Precision & 0.856 & 0.847 & 0.838 & 0.739 & 0.868 & 0.956 & 0.744 & 0.970 \\
& Recall & 0.950 & 0.940 & 0.933 & 0.831 & 0.951 & 0.972 & 0.870 & 0.970 \\
\hline
\multirow{3}{*}{AOES (RoBERTa) } & F1 & \textbf{0.905} & \textbf{0.892} & \textbf{0.913} & 0.659 & 0.866 & 0.947 & 0.746 & 0.921 \\
& Precision & 0.864 & 0.841 & 0.877 & 0.600 & 0.808 & 0.930 & 0.664 & 0.904 \\
& Recall & 0.95 & 0.95 & 0.952 & 0.731 & 0.933 & 0.964 & 0.85 & 0.94 \\
\hline
\end{tabular}
\caption{Off-Topic Detection Quantitative Comparison Results on ASAP-AES Dataset}
\label{tab:asap_offtopic}
\end{table*}

\begin{table*}
\scriptsize
\centering
\begin{tabular}{|c|c|c|c|c|c|c|c|c|c|c|} 
\hline
\multirow{2}{*}{\textbf{Model}} & \multirow{2}{*}{\textbf{Metric}} & \multicolumn{9}{c|}{\textbf{Prompt ID }} \\ 
\cline{3-11}
 &  & \textbf{1} & \textbf{2} & \textbf{3} & \textbf{4} & \textbf{5} & \textbf{6} & \textbf{7} & \textbf{8} & \textbf{9} \\ 
\hline
\multirow{3}{*}{Baseline 1}  & F1 & 0.602 & 0.920 & 0.874 & 0.847 & 0.758 & 0.645 & 0.613 & 0.787 & 0.721 \\ 
 & Precision & 0.985 & 0.991 & 0.973 & 0.998 & 0.998 & 0.998 & 0.998 & 0.998 & 0.976 \\ 
 & Recall & 0.433 & 0.858 & 0.794 & 0.734 & 0.610 & 0.476 & 0.442 & 0.648 & 0.571 \\  
\hline
\multirow{3}{*}{Baseline 2} & F1  & 0.771 &  0.871 &  0.874 &  0.868 &  0.807 &  0.676 &  0.703 &  0.800 &  0.807 \\ 
 & Precision & 0.985 & 0.91 & 0.93 & 0.964 & 0.953 & 0.978 & 0.981 & 0.976 & 0.956 \\ 
 & Recall & 0.633 & 0.835 & 0.824 & 0.789 & 0.700 & 0.517 & 0.548 & 0.678 & 0.698 \\  
\hline
\multirow{3}{*}{Louis \textit{et al.}}  & F1 & 0.699 & 0.690 & 0.527 & 0.664 & 0.782 & 0.710 & 0.672 & 0.604 & 0.743 \\ 
 & Precision & 0.718 & 0.687 & 0.525 & 0.671 & 0.839 & 0.720 & 0.728 & 0.644 & 0.805 \\ 
 & Recall & 0.68 & 0.693 & 0.529 & 0.657 & 0.732 & 0.701 & 0.624 & 0.570 & 0.690 \\ 
\hline
\multirow{3}{*}{Shahzad \textit{et al.}}  & F1 & 0.826 & 0.416 & 0.200 & 0.728 & 0.727 & 0.630 & 0.637 & 0.263 & 0.621 \\ 
 & Precision & 0.982 & 0.783 & 0.667 & 0.906 & 0.954 & 0.841 & 0.959 & 0.788 & 0.990 \\ 
 & Recall & 0.713 & 0.283 & 0.118 & 0.608 & 0.587 & 0.503 & 0.477 & 0.158 & 0.452 \\ 
\hline
\multirow{3}{*}{ \textit{Raina et al.}}  & F1 & 0.816 & 0.719 & 0.805 & 0.739 & 0.836 & 0.751 & 0.831 & 0.817 & \textbf{0.898} \\ 
 & Precision & 0.845 & 0.754 & 0.812 & 0.847 & 0.868 & 0.843 & 0.959 & 0.85 & 0.910 \\ 
 & Recall & 0.789 & 0.687 & 0.798 & 0.656 & 0.807 & 0.678 & 0.734 & 0.787 & 0.886 \\ 
\hline
\multirow{3}{*}{AOES (BERT)}  & F1 & \textbf{0.899} & \textbf{0.929} & \textbf{0.894} & 0.919 & \textbf{0.887} & \textbf{0.869} & \textbf{0.888} & 0.884 & 0.804 \\ 
 & Precision & 0.911 & 0.929 & 0.902 & 0.928 & 0.913 & 0.881 & 0.914 & 0.905 & 0.851 \\ 
 & Recall & 0.887 & 0.929 & 0.886 & 0.909 & 0.864 & 0.857 & 0.863 & 0.864 & 0.762 \\ 
\hline
\multirow{3}{*}{AOES (RoBERTa)}  & F1 & 0.854 & 0.889 & 0.875 & \textbf{0.947} & 0.886 & 0.845 & 0.863 & \textbf{0.905} & 0.831 \\ 
 & Precision & 0.869 & 0.896 & 0.875 & 0.951 & 0.919 & 0.854 & 0.896 & 0.919 & 0.87 \\ 
 & Recall & 0.84 & 0.882 & 0.875 & 0.944 & 0.854 & 0.837 & 0.832 & 0.891 & 0.795 \\ 
\hline
\end{tabular}
\caption{Off-Topic Detection Quantitative Comparison Results on PsyW-Essay Dataset}
\label{tab:PsyW-Essay_offtopic}
\end{table*}

\textbf{Off Topic Dataset Creation : } We sampled off-topic essays from each prompt excluding the ones that are not part of the training dataset to measure the performance of off-topic detection. All prompts used in the off-topic test set are carefully checked so that they are different from the other essay prompts. To rule out the possibility that the model overfits to the training off-topic data, we sampled the off-topic essay for the test set from the prompts different from the ones in the training off-topic data.
For each prompt in the ASAP-AES dataset, we randomly selected three other prompts, sampled 200 data for the off-topic train split, and sampled 100 samples from the rest four prompts for the off-topic test split.
In the case of the PsyW-Essay Dataset, we randomly selected 4 prompts and sampled 150 data for the off-topic train split and collected 100 samples from the remaining prompts for the off-topic test set.
We created an off-topic training set to train one of our baselines for comparison with our approach. An off-topic training split is not used in our proposed method. Only the off-topic test set is used evaluation purposes.

\subsection{Evaluation Metric}
\textbf{Off Topic Evaluation Metrics : }
Based on the previous work \citet{wang2019automatic,yoon2017off,shahzad2022computerization}, we used Precision, Recall, and F1 score for the evaluation of off-topic essay detection. \newline
\textbf{Essay Scoring Evaluation Metrics : }
As mentioned in previous work~\citet{wang2022use,yang2020enhancing}, Quadratic Weighted Kappa (QWK) is used as the essay scoring metric, which measures the agreement between estimated scores and ground truth scores. We are also using Pearson Correlation Coefficient to evaluate the degree of strength, and direction of association between predicted essay scores and graded essay scores. 
For on-topic essay score estimation evaluation, a higher value of Quadratic Weighted Kappa (QWK) and Pearson Correlation Coefficient indicates the higher performance of the score estimation model.

\subsection{Training and Inference Details}
The proposed AOES model is trained on the on-topic training dataset for essay scoring. We have trained AOES for 20 epochs with 16 batch size and also used a learning rate of 5e-4 with 500 warm-up steps. We chose $\lambda = 0.6$ since it performed the best across all datasets in our experiment. We have used the same hyperparameters across all datasets. After training, the mean and covariance matrices of all hidden layers for the on-topic training data from the corresponding prompt are saved. At the time of inference, the Mahalanobis distance score of the essay text from test data is calculated using the previously saved mean feature vector and covariance matrix of the corresponding training set, and that score is used as the measure of the off-topic detection. The proposed AOES model is evaluated on a test set that consists of an on-topic test set and an off-topic test set. 

\subsection{Baseline}
We have created two supervised method-based baselines, Baseline-1~\label{baseline-1} and Baseline-2~\label{baseline-2} where both follow the same backbone model architecture, but both have their own different training and inference strategies.

\textbf{Baseline 1:} \label{baseline-1} We implemented the BERT model and  pooled output of the model was fed into the linear layer with one output for the regression task, which intends to minimize the mean squared error loss, $\mathcal{L}_{MSE}$ while training. We trained the model on the regression task using both the on-topic training dataset and the off-topic training dataset, where samples from the off-topic dataset are rated as zero grade.
During inference, the predicted essay score is used to detect the off-topic text by applying a threshold value directly, instead of the Mahalanobis distance score. The reason behind using this supervised baseline method is to justify the performance, and robustness of our proposed unsupervised method by utilizing the unique loss function during training and the Mahalanobis distance score for off-topic detection.

\textbf{Baseline 2:} \label{baseline-2} We developed a multi-task learning-based baseline model with two distinct branches in the final layer—one for estimating essay scores and the other for off-topic detection. The model produces two values in response to the input essay: the essay score and topic class. Here, we jointly train the model to score essays and classify essays as on-topic or off-topic, by optimizing a joint multi-task loss function,$\mathcal{L}_{multi-task} = = \mathcal{L}_{MSE} + \mathcal{L}_{BCE}$ where essay scoring logit and topic classification logit optimize mean squared error loss, $\mathcal{L}_{MSE}$ and binary cross-entropy loss, $\mathcal{L}_{BCE}$ respectively. The goal of adopting a multi-task learning-based baseline technique is to learn the representations between the two tasks in order to improve generalization and performance on both. 

\subsection{Results}

\subsubsection{Essay Scoring Performance}
Table~\ref{tab:PsyW_asap_ontopic} shows the results of the essay scoring of the on-topic ASAP-AES dataset and PsyW-Essay dataset. The proposed method shows relatively good results on the QWK score and correlation on each dataset. 
\begin{table*}
\scriptsize
\centering
\begin{tabular}{|c|c|c|c|c|c|c|c|c|c|c|c|} 
\hline
\multirow{2}{*}{\textbf{Dataset}}     & \multirow{2}{*}{\textbf{Model}}               & \multirow{2}{*}{\textbf{Metric}} & \multicolumn{9}{c|}{\textbf{Prompt ID }}                                                                                                                \\ 
\cline{4-12}
                                      &                                               &                                  & \textbf{1}     & \textbf{2}     & \textbf{3}     & \textbf{4}     & \textbf{5}     & \textbf{6}     & \textbf{7}     & \textbf{8}     & \textbf{9}      \\ 
\hline
\multirow{6}{*}{PsyW-Essay } & \multirow{2}{*}{AOES (without TRM)}  & QWK                              & 0.698          & 0.720          & 0.723          & 0.796          & 0.733          & 0.713          & 0.752          & 0.710          & 0.709           \\
                                      &                                               & Correlation                      & 0.781          & 0.786          & 0.873          & 0.890          & 0.791          & 0.753          & 0.876          & 0.799          & 0.786           \\ 
\cline{2-12}
                                      & \multirow{2}{*}{AOES (with L2 Loss)} & QWK                              & 0.698          & 0.723          & \textbf{0.742} & 0.770          & 0.737          & 0.719          & 0.795          & 0.711          & 0.699           \\
                                      &                                               & Correlation                      & 0.787          & 0.784          & \textbf{0.903} & 0.903          & 0.796          &\textbf{0.832}          & \textbf{0.881} & 0.801          & 0.805           \\ 
\cline{2-12}
                                      & \multirow{2}{*}{AOES (BERT)}                & QWK                              & \textbf{0.722} & \textbf{0.734} & 0.733 & \textbf{0.826} & \textbf{0.742} & \textbf{0.720} & \textbf{0.807} & \textbf{0.728} & \textbf{0.763}  \\
                                      &                                               & Correlation                      & \textbf{0.809} & \textbf{0.797} & 0.899 & \textbf{0.906} & \textbf{0.814} & 0.822 & 0.878  & \textbf{0.803} & \textbf{0.824}  \\
\hline
\hline
\multirow{6}{*}{ASAP-AES} & \multirow{2}{*}{AOES (without TRM)}  & QWK                              & 0.784          & 0.647          & 0.611           & 0.784          & 0.741           & 0.765          & 0.839          & 0.749       & -     \\
                                   &                                               & Correlation                      & 0.832          & 0.731          & 0.685           & 0.822          & 0.817           & 0.828          & 0.855          & 0.737        & -    \\ 
\cline{2-12}
                                   & \multirow{2}{*}{AOES (with L2 Loss)} & QWK                              & 0.791  & \textbf{0.678} & 0.596 & 0.748  & 0.768  & \textbf{0.792} & 0.842  & 0.752   & -   \\
                                   &                                               & Correlation                      & 0.821  & \textbf{0.749}  & 0.645  & 0.819  & \textbf{0.828}  & 0.837  & 0.859 & 0.775  & -          \\ 
\cline{2-12}
                                   & \multirow{2}{*}{AOES (BERT)}               & QWK                              & \textbf{0.793} & 0.661  &   \textbf{0.667} & \textbf{0.789} & \textbf{0.782} & 0.787 & \textbf{0.845} & \textbf{0.754} & -  \\
                                   &                                               & Correlation                      & \textbf{0.834} & 0.741  & \textbf{0.695}  & \textbf{0.836} & 0.825  & \textbf{0.842} & \textbf{0.863} & \textbf{0.789} & -  \\
\hline
\end{tabular}
\caption{On-Topic Automatic Essay Scoring Results of the Ablation Study on Both ASAP-AES and PsyW-Essay Datasets}
\label{tab:PsyW_asap_ontopic_ablation}
\end{table*}

\subsubsection{Off Topic Performance}
Off-topic detection performance on ASAP-AES off-topic test set and PsyW-Essay off-topic test set are shown in Table~\ref{tab:asap_offtopic} and Table~\ref{tab:PsyW-Essay_offtopic} respectively. For baseline and the proposed method, the reported results are on an equal error rate threshold which means precision and recall have the same importance during off-topic classification. As from Table~\ref{tab:asap_offtopic} and Table~\ref{tab:PsyW-Essay_offtopic}, the proposed method shows a significant improvement in F1 score with respect to baseline. As Baseline is a supervised technique, its success is reliant on the off-topic training data, which causes it to succeed on certain prompts while failing on others.

\subsubsection{Performance of Different Model Architectures}
We also experimented with the proposed TRM module with the RoBERTa model as the backbone to validate that the TRM module can be attached to different pre-trained models. Both on-topic and off-topic performance of this RoBERTa based model are reported in Table~\ref{tab:asap_offtopic}, Table~\ref{tab:PsyW-Essay_offtopic} and Table~\ref{tab:PsyW_asap_ontopic}.

\subsubsection{Quantitative Comparison}
We evaluate our proposed method with two previously proposed methods. The first method~\citet{louis2010off} suggested a technique that compares the TF-IDF similarity between the prompt and the given response. As our proposed method is an unsupervised method, we chose Louis et al as it was the only available unsupervised approach for off-topic detection. The second method~\citet{shahzad2022computerization} proposed a solution that uses a random forest classifier and concatenated feature representations from the Word Mover’s Distance ~\citet{kusner2015word}, IDF-weighted word embedding similarity, and the average embedding similarity of the Word2vec embedding ~\citet{mikolov2013efficient}. The last method uses a combined Hierarchical attention-based topic model (HATM) and Similarity Grid model (SGM) for off-topic essay detection. We particularly chose the following supervised methods, Raina et al. and Shahzad et al. as these are the latest works in off-topic detection domain and they also compare different state-of-the-art supervised methods. Performance of these previously proposed supervised methods is reported in Table~\ref{tab:asap_offtopic} and Table~\ref{tab:PsyW-Essay_offtopic} for ASAP-AES and PsyW-Essay datasets respectively. As we can see from these tables, our approach, based on the Mahalanobis distance score, outperforms the earlier works by a large margin.

\subsubsection{Qualitative Analysis}
\begin{figure}[]
  \centering
  \includegraphics[scale=0.57]{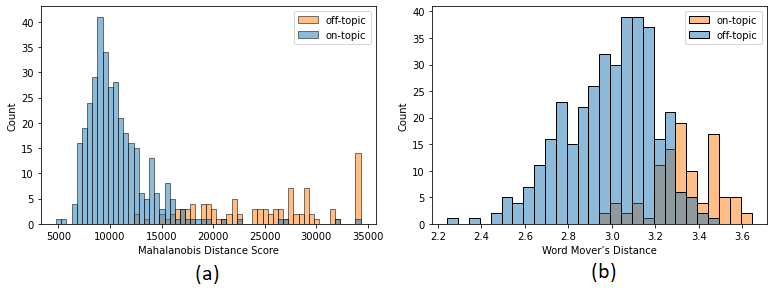}
  \caption{Histogram of Detection Scores using Various Methods on a single prompt from ASAP-AES dataset. Here, (a) AOES (b) Word Mover Distance~\cite{shahzad2022computerization}}
  \label{fig:aes_qual}
\end{figure}

\begin{figure}[]
  \centering
  \includegraphics[scale=0.57]{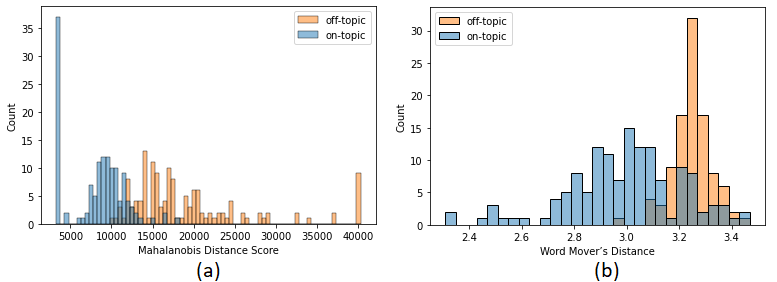}
  \caption{Histogram of Detection Scores using Various Methods on a single prompt from PsyW-Essay dataset. Here, (a) AOES (b) Word Mover Distance~\cite{shahzad2022computerization} }
  \label{fig:shl_qual}
\end{figure}

We provide a quantitative analysis by visualizing histogram plots of detection scores for on-topic and off-topic data. As an off-topic detection score, we use the Mahalanobis distance score for the proposed AOES model and word mover distance from previous works~\cite{shahzad2022computerization,yoon2017off}. Histogram plots of both types of distance are shown in Figure~\ref{fig:aes_qual} for ASAP-AES and in Figure~\ref{fig:shl_qual} for the PsyW-Essay dataset. From plots of respective datasets, it is prominent that AOES significantly reduces the overlap between on-topic and off-topic in the first subfigure compared to the other subfigure. 

\subsection{Ablation Study}
We examine the effects of our two unique components, the TRM Layer and proposed loss function, on the performance.
\subsubsection{Importance of TRM Layer}
To verify the contribution of the proposed TRM layer in the AOES model, we used a similar BERT regression model without the TRM layer. This model is trained in the same unsupervised setting by applying the same data and training configuration as the proposed unsupervised method. Mahalanobis distance score is also used for off-topic detection. On-topic performance results of this model are shown in Table~\ref{tab:PsyW_asap_ontopic} for both ASAP-AES and PsyW Essay Dataset. Similarly, off-topic performance results of this model is also reported in Table~\ref{tab:PsyW-Essay_offtopic}. As seen in these tables for both datasets, the TRM layer is essential for improving the overall performance of the AOES model. 
\subsubsection{Effect of Proposed ${L}_{Topic}$ Loss function}
As discussed in Section~\ref{training_loss_func}, $\mathcal{L}_{Topic}$ is used to incorporate extra regularization by confining the output value of the topic branch, $Y_t$ between 0 and 1. This property also can be achieved by using L2 loss instead of the proposed loss function. We train a AOES model with L2 loss as ${L}_{Topic} = \frac{1}{N}\sum_{i=1}^{N}||1 - Y^i_t||^2$, to demonstrate the significance of proposed topic regularization loss. 
Both on-topic and off-topic performance results of this study are reported in Table~\ref{tab:PsyW_asap_ontopic_ablation} and Table~\ref{tab:PsyW_asap_offtopic_ablation} for ASAP-AES dataset and PsyW-Essay dataset.
The reported results from these tables show that the suggested loss ${L}_{Topic}$ performs best for the topic regularization loss.

\subsection{Performance on Adversarial Sample Detection}
We experimented with several perturbation techniques discussed in previous studies~\citet{kabra2022evaluation,ding2020don} to generate adversarial examples, vulnerable to current AES systems. Here, we use the suggested method to detect these adversarial input samples to check the robustness of our model against these perturbations.
More details on these perturbation techniques are given below.
\begin{enumerate}
    \item \textit{AddSpeech} - As per \citet{kabra2022evaluation} study, we extracted speech from the famous leader into the test responses and created the off-topic response by adding these irrelevant speech sentences to the test response. According to our qualitative study of experiment results, the number of sentences added in the AddSpeech adversarial transformation affects the overall detection scores. We observed that off-topic responses which have a very high number of speech sentences are hard to detect using our framework.

    \item \textit{BabelGenerate} - We use  B.S. Essay Language Generator (BABEL)~\citet{perelman2020babel} to generate gibberish samples from some keywords which we manually created for each prompt. We manually created these keywords for each prompt by looking at important and relevant words found in on-topic samples of the corresponding prompt.
    These keywords are used to generate an incoherent, meaningless passage containing a concoction of obscure words and keywords concatenated together.
    \item \textit{RepeatSent} - In order to make responses longer without going off-topic and to create coherent paragraphs, students often deliberately repeat sentences or particular keywords. To create such responses, we randomly sample sentences and repeat them an arbitrary number of times and add them back to the response. According to our qualitative study of experiment results, the number of times sentences are repeated in the PsyW dataset is comparatively low, making it relatively difficult to classify as off-topic, as few repeated sentences imply adversarial and the input essays are very similar. 

    \item \textit{ReplaceSents} - Another common strategy to bluff an exam is to write something unrelated in the middle of the essay, while the initial and final parts are on topic. We simulate this by substituting other off-topic responses only for the body paragraphs of the responses, keeping the first and last sentences on-topic.  

    \item \textit{GPTGenerate} - It is possible to generate essays through generative models like \footnote[1]{https://huggingface.co/gpt2}GPT-2 and \footnote[2]{https://huggingface.co/EleutherAI/gpt-neo-2.7B}GPT-Neo-2.7b that may appear to be well-written and on-topic but are actually off-topic and irrelevant. This technique can be used to bluff an exam by submitting a seemingly coherent essay that does not actually answer the question. We generated an essay from the GPT-based models on the given prompts to verify that our system can detect AI-generated coherent off-topic essays.
\end{enumerate}{}

\begin{table}
\centering
\scriptsize
\resizebox{7.7cm}{!}{
\begin{tabular}{|c|c|c|c|c|} 
\hline
\textbf{Dataset}      & \begin{tabular}[c]{@{}c@{}}\textbf{Type of }\\\textbf{Adversaries}\end{tabular} & \textbf{F1} & \textbf{Precision~} & \textbf{Recall}  \\ 
\hline
\multirow{6}{*}{PsyW}  & \textit{AddSpeech}                    & ~     0.748      & ~         0.754          & ~       0.741         \\ 
\cline{2-5}
                      & \textit{BabelGenerate}                   & ~      0.931     & ~         	0.931          & ~        0.931        \\ 
\cline{2-5}
                      & \textit{RepeatSent}                    & ~     0.853      & ~         0.845          & ~         0.861       \\ 

\cline{2-5}
                      & \textit{ReplaceSents}                       & ~      0.909     & ~        0.913           & ~        0.905        \\ 
\cline{2-5}
                      & \textit{GPTGenerate[GPT-2]}                       & ~      0.976     & ~        0.979           & ~        0.972        \\ 
\cline{2-5}
                      & \textit{GPTGenerate[GPT-Neo]}                       & ~      0.97     & ~        0.972           & ~  0.968              \\ 
\hline
\multirow{6}{*}{ASAP} & \textit{AddSpeech}                    & ~     0.752      & ~         0.752          & ~        0.752        \\ 
\cline{2-5}
                      & \textit{BabelGenerate}                   & ~      0.997     & ~          0.993         & ~        0.998        \\ 
\cline{2-5}
                      & \textit{RepeatSent}                    & ~      0.993     & ~         0.994          & ~      0.990          \\ 
\cline{2-5}
                      & \textit{ReplaceSents}                       & ~      0.914     & ~        0.917           & ~       0.910         \\
\cline{2-5}
                      & \textit{GPTGenerate[GPT-2]}                       & ~      0.950     & ~        0.954           & ~  0.946              \\ 
\cline{2-5}
                      & \textit{GPTGenerate[GPT-Neo]}                       & ~      0.905     & ~        0.912           & ~  0.898              \\ 
\hline
\end{tabular}
}
\caption{Adversarial Sample Detection Results on Different Perturbation Techniques on Both Datasets}
\label{tab:adv_results}
\end{table}
According to Table~\ref{tab:adv_results}, it is prominent that our proposed method can effectively distinguish these adversarial responses. As the babel-generated essays based on keywords are irrelevant and incoherent, Mahalanobis distance can effectively distinguish these generated responses. Similarly, responses with unrelated content in body paragraphs are also distinguished effectively. 
 

\section{Conclusion}
This paper proposes a joint transformer-based model, using only on-topic essay examples to estimate on-topic essay scores and detect off-topic essay responses for the Automated Essay Scoring (AES) System in an open-world setting. Our proposed TRM layer is used as a drop-in replacement for the last layer in the transformer-based AES model, providing a low-cost approach with significant improvement. For off-topic detection, we use the Mahalanobis distance score, which greatly enhances the detection ability and lowers computational costs. We have also shown on two datasets that our method can detect adversarial samples effectively without compromising on-topic performance.
In the future, we will investigate more with long-formers and other methods to effectively encode long essay corpora in vector space to improve essay scoring and off-topic performance.

\section{Ethical Considerations and Limitations}

We haven't studied social bias or any other adverse impact category because all user data—like gender, race, and other details—that is necessary to identify any kind of social bias was missing from both essay datasets. Other than that, there is no social bias in the question or prompt that was used in both datasets. A group of I/O psychologists created our in-house dataset in this way to prevent biases of that kind. Furthermore, although various perturbation scenarios are taken into account for evaluation, there might be an additional means of deceiving the AES system that we are not aware of.







\section{References}\label{sec:reference}

\bibliographystyle{lrec-coling2024-natbib}
\bibliography{lrec-coling2024-example}

\end{document}